\documentclass[letterpaper, 10 pt, conference]{ieeeconf}  % Comment this line out if you need a4paper

\IEEEoverridecommandlockouts                              % This command is only needed if 
\usepackage{cite}
\usepackage{amsmath,amssymb,amsfonts}
\usepackage{algorithmic}
\usepackage{graphicx}
\usepackage{textcomp}
\usepackage{xcolor}
\usepackage{caption}
\usepackage{subcaption}
\usepackage{multirow}
\usepackage{soul}
\usepackage{balance}
\usepackage[T1]{fontenc}

\title{\LARGE \bf
Hierarchical Policy for Non-prehensile Multi-object Rearrangement with Deep Reinforcement Learning and Monte Carlo Tree Search
}

\author{Fan Bai, Fei Meng, Jianbang Liu, Jiankun Wang and Max Q.-H. Meng$^*$, \textit{Fellow, IEEE}% <-this % stops a space
\thanks{$^*$Corresponding author.}% <-this % stops a space
\thanks{Fan Bai, Fei Meng and Jianbang Liu are with the Department of Electronic Engineering, The Chinese University of Hong Kong, Hong Kong. {\tt\small \{fanbai, feimeng, henryliu \}@link.cuhk.edu.hk}. Jiankun Wang and Max Q.-H. Meng are with the Department of Electronic and Electrical Engineering, Southern University of Science and Technology, Shenzhen, China {\tt\small wangjk@sustech.edu.cn}. Max Q.-H. Meng is on leave from the Department of Electronic Engineering, The Chinese University of Hong Kong, Hong Kong, and also with the Shenzhen Research Institute of the Chinese University of Hong Kong, Shenzhen, China. {\tt\small max.meng@ieee.org}.}%
}

\begin{document}

\maketitle
\thispagestyle{empty}
\pagestyle{empty}

%%%%%%%%%%%%%%%%%%%%%%%%%%%%%%%%%%%%%%%%%%%%%%%%%%%%%%%%%%%%%%%%%%%%%%%%%%%%%%%%
\begin{abstract}

Non-prehensile multi-object rearrangement is a robotic task of planning feasible paths and transferring multiple objects to their predefined target poses without grasping. 
It needs to consider how each object reaches the target and the order of object movement, which significantly deepens the complexity of the problem.
To address these challenges, we propose a hierarchical policy to divide and conquer for non-prehensile multi-object rearrangement. In the high-level policy, guided by a designed policy network, the Monte Carlo Tree Search efficiently searches for the optimal rearrangement sequence among multiple objects, which benefits from imitation and reinforcement.  
In the low-level policy, the robot plans the paths according to the order of path primitives and manipulates the objects to approach the goal poses one by one. 
We verify through experiments that the proposed method can achieve a higher success rate, fewer steps, and shorter path length compared with the state-of-the-art.

\end{abstract}

%%%%%%%%%%%%%%%%%%%%%%%%%%%%%%%%%%%%%%%%%%%%%%%%%%%%%%%%%%%%%%%%%%%%%%%%%%%%%%%%

\section{Introduction}

The non-prehensile multi-object (NPMO) rearrangement defines a task that the robot manipulates multiple objects into predefined target poses in an unstructured environment. 
Generally, the task space is limited to a table, and the robot only takes non-prehensile actions such as pushing \cite{lynch1996stable}, sliding \cite{vose2007vibration,maeda2004motion} and batting \cite{anderson1988robot}, without pick-and-place actions. 
Since the robot does not pick up the object, the object will not leave the table with non-prehensile actions. Therefore, the safety of the manipulation is guaranteed. People can apply it to special scenes, such as rearranging objects that are difficult to grasp, large, heavy, or easily damaged.

The previous non-prehensile rearrangement work mainly focuses on how to plan a collision-free path for one object \cite{haustein2015kinodynamic,haustein2018non,king2015nonprehensile,king2016rearrangement}, which only solves the problem of \textit{how to go}. 
However, for the NPMO rearrangement problem, the robot not only needs to plan a rearrangement path for each object but also needs to determine \textit{what order to follow}, which is known as an NP-hard problem.

\begin{figure}[!ht]
    \centerline{\includegraphics[width=0.45\textwidth]{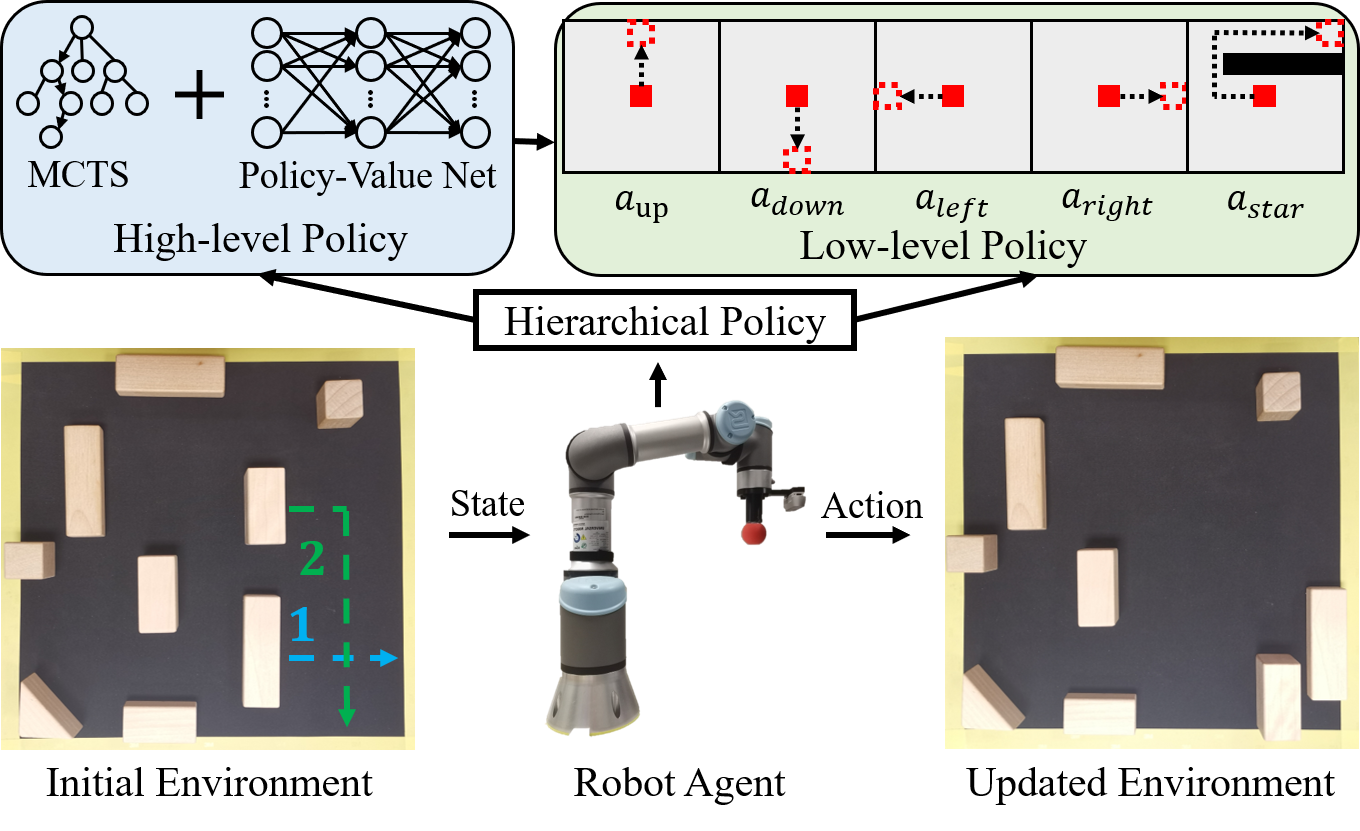}}
    \caption{Illustration of the hierarchical policy process. Given the initial and target state, the robot first determines which object and path primitive action will be selected via high-level MCTS policy. Then the robot manipulates this object to achieve the path primitives using the low-level policy.}
    \label{fig1}
    \vspace{-0.5cm} 
\end{figure}

Recently, Song \textit{et al.}\cite{song2019multi} proposed the NPMO rearrangement with the Monte Carlo tree search (MCTS) method, which is capable of sorting and classifying large numbers of objects reliably. Huang \textit{et al.}\cite{huang2019large} showed that using an iterated local search technique can search the push policy sufficiently.
However, on the one hand, they only concern the rearrangement of multi objects by region rather than emphasizing the path blocking between each other. For example, in Fig. \ref{fig1}, if the robot first moves the object following the green path 2, it will block the desired blue path 1 of another object, thus increasing the total length of object paths. On the other hand, as the number of objects increases, the search algorithm may fail and its search steps and length of move action sequence will increase sharply. 
In our task, it is a challenging problem to taking into account \textit{how to go} and \textit{what order to follow}, i.e., it has to plan a path for each object while considering the interaction between them. Yuan \textit{et al.} \cite{yuan2018rearrangement} leveraged end-to-end deep reinforcement learning to generate control signals from images, which increased the difficulty of training and the depth of decision-making. It is easy to fall into local optimal values and challenging to train for complex tasks under sparse rewards. If traditional path planning algorithms, e.g., A*, RRT, etc., are used for the NPMO rearrangement, it only ensures that the path of a single object is reasonable but cannot consider the mutual influence of the paths between multiple objects.  

This work proposes a hierarchical policy to address the NPMO rearrangement, which improves the search speed and enhances long-term decision-making capabilities. The flow chart of our method is shown in Fig. \ref{fig1}. In the low-level policy, the robot directly plans the paths using the path primitives, including some basic motion sequences used to push the object to achieve the goal steadily. Different from the single-step discrete action \cite{yuan2018rearrangement,song2019multi}, the designed path primitive reduces the depth and width of the search tree. In the high-level policy, the robot can orchestrate these path primitives to obtain the optimal rearrangement policy of multiple objects through the strong long-term decision-making ability of MCTS guided by the policy network. 

The contributions of this work are summarized as follows:

1) We model the NPMO rearrangement task and solve it with a hierarchical policy;

2) We propose a high-level MCTS policy accelerated by the policy network trained with imitation and reinforcement;

3) We design a low-level policy to control the robot to achieve path primitives execution.

\section{Related Work}
\textbf{The NPMO Rearrangement:} Unlike other rearrangement tasks, the robot needs to rearrange the objects from the initial position to the target position without grasping in the NPMO rearrangement task. Since grasping is forbidden, the robot generally performs rearrangement by pushing, rolling, and dragging\cite{ruggiero2018nonprehensile}, etc. Although these operations increase the task's difficulty, they attract more robot researchers to solve the NPMO rearrangement task due to the safety and practical significance of special scenes. The current NPMO rearrangement of objects can be divided into two parts: \textit{how to go} and \textit{what order to follow}.

\textit{How to go} has been studied maturely. \cite{haustein2015kinodynamic,haustein2018non,king2015nonprehensile,yuan2018rearrangement,YUAN2019119,huang2021visual}  proposed a series of path planning, deep reinforcement learning, and search algorithms to solve it. For multiple objects, the question of \textit{what order to follow} is more critical. Song \textit{et al.}\cite{song2019multi} proposed to employ MCTS to do the sorting task, which separated objects belonging to different classes into homogenous distinct clusters. Even if they explored the rearrangement of multiple objects, their actions are a single step instead of a path sequence. The width and depth of the search increase, significantly reducing the algorithm's efficiency, and it is difficult to focus on the order of objects.

\textbf{Sequential Decision Making:} We formulate the NPMO rearrangement problem as a sequential decision problem, which is further represented as a Markov decision process. In sequential decision making, the action of the previous step will affect the next step. Many problems can be represented in this form, such as game playing\cite{silver2016mastering,zha2021douzero,won2020adaptive}, autonomous driving\cite{chen2019model,osinski2020simulation,pan2017virtual}, robot control\cite{gu2017deep,sadeghi2017sim2real}, recommended system\cite{li2010contextual} and trading\cite{deng2016deep}. Inspired by this work, we also adopt the classic architecture in which the policy network guides the MCTS. The scene rearrangement problem \cite{wang2020scene} is similar to our work. But there are still many differences. First, to use expert-labeled action data, we use a policy network that can be directly used for imitation learning instead of Deep Q Network in \cite{wang2020scene}. In the problem definition, the robot NPMO rearrangement problem is more complicated. It is not only necessary to consider the sequential decision, the non-prehensile manipulation, but also needs to be combined with robot control. And our path primitive requires the robot to plan and implement the paths.

\textbf{Imitation Learning:} In real tasks, human experts can often perform tasks well, but it is difficult for agents to do it. To make the agent reach the expert level quickly, we leverage the prior information and data of the expert to do imitation learning. Among them, the method of directly matching the state and action distribution of the model with the distribution of expert data is called behavior cloning, which has an excellent performance in decision-making, such as autonomous driving\cite{bojarski2016end} and game playing\cite{silver2016mastering}. But in our problem, it is difficult to achieve the best performance using only imitation learning because it is difficult for humans to make correct decisions as the number of objects increases. Thus, in our work, the policy network imitates the expert and is further improved through reinforcement learning.

\section{Problem Definition}
\subsection{Task and Assumptions}

\begin{figure}[!ht]
    \centerline{\includegraphics[width=0.4\textwidth]{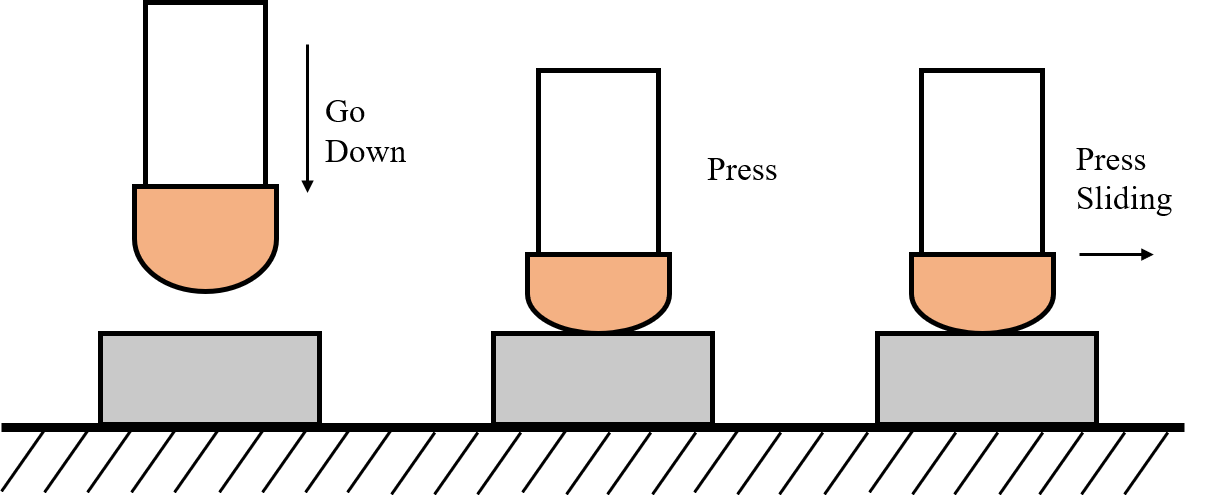}}
    \caption{Non-prehensile manipulation tool and execution way.}
    \label{fig2_tool}
\vspace{-0.2cm} 
\end{figure}

This work assumes that a robot is equipped with a non-prehensile manipulation tool shown in Fig. \ref{fig2_tool}, which can take non-prehensile actions such as pushing and cannot take pick-and-place actions.

\vspace{-0.2cm} 

Our task is to find sequences of actions and control the robot to push all the objects from the initial poses to the target poses while avoiding collisions. This process can be regarded as a quasi-static process which is assumed that any inertial forces can be neglected due to frictional forces.

\begin{figure*}[htbp]
    \centerline{\includegraphics[width=0.9\textwidth]{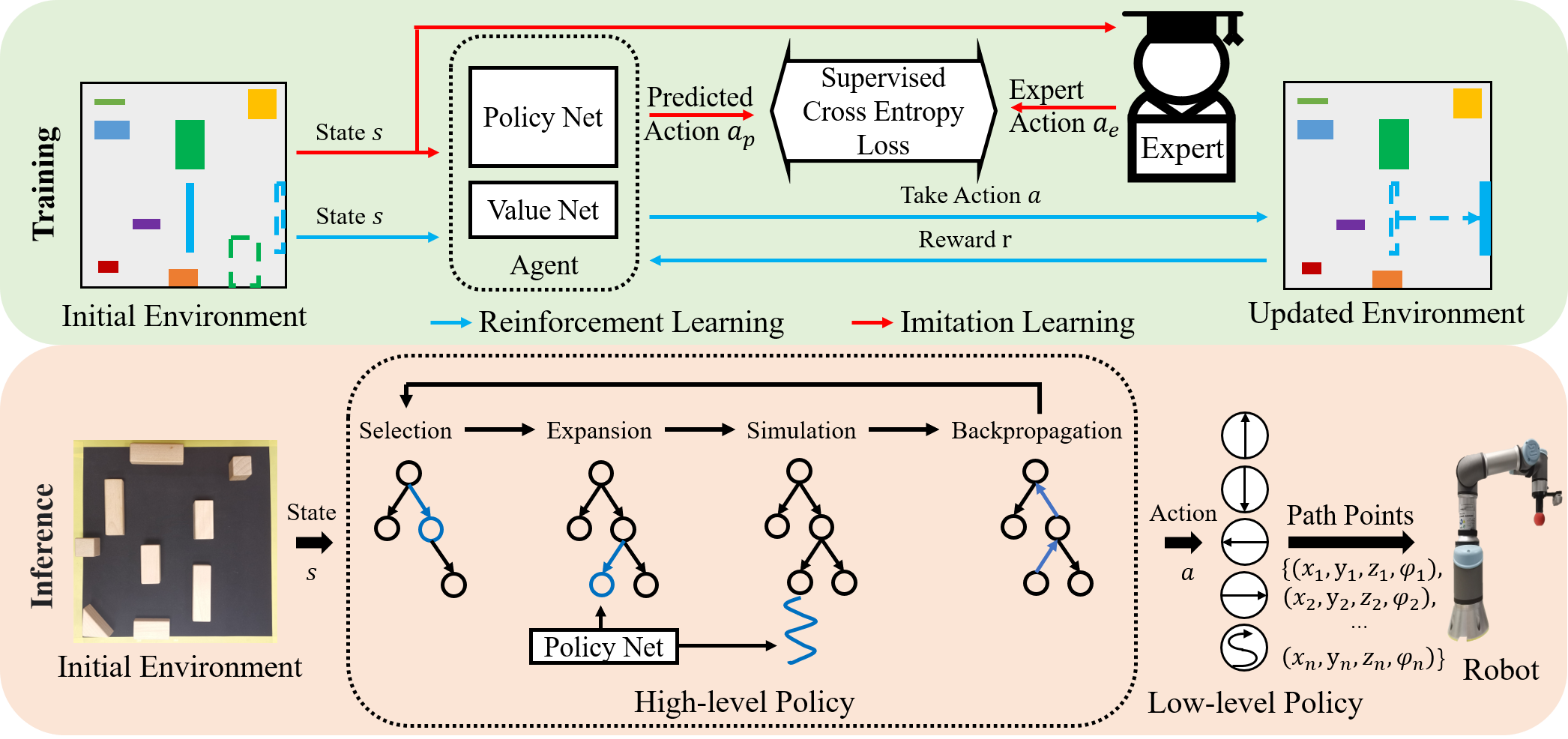}}
    \caption{Illustration of the training and inference process.}
    \label{fig3_pipline}
\vspace{-0.5cm} 
\end{figure*}

\subsection{Problem Formulation}
In the NPMO rearrangement task, we can consider it as a sequential decision-making problem, where the robot $R$ perceives one observation of an initial state $s_I$ and a target state $s_T$. The aim is to take the collision-free path from $s_I$ to $s_T$ with an action sequence $A = (a_1,a_2,...,a_n)$, where $a_i \in A$,  $i=1,2,...,n$, and $A$ represents the action space. Simultaneously, we maximize the accumulated reward and minimize the move length $L_t$ and steps of multiple objects.

Formally, the objective function representing an accumulated reward can be written as
\begin{equation}
    \mathop{\arg\max}_{A} \sum_{t=1}^{n} {r(s_t,a_t)}
\end{equation}
where $s_1 = s_I$ and $s_{n+1} = s_T$. $(S,A)$ $\rightarrow$ $R$ donates a reward function, given $s_t \in S$ and $a_t \in A$, $r(s_t,a_t) \in R$ is a reward in each time step $t$.

\textbf{Observation State Space:} The observation of the state $s_t \in S$ is a $M\times M\times (2N+1)$ matrix, $M$ means we discretize the image into a $M \times M$ grid, $2N+1$ means it can represent up to $N$ movable objects and 1 channel immovable object, and the initial and target pose of each object is masked by one-hot encoding.

\textbf{Move Action Space:} In order to reduce the width and depth of the search, we chose the path primitives as the action space $\{a_{up}, a_{down}, a_{left}, a_{right}, a_{star}\} \in A$. These low-level path primitives mean moving along the up, down, left, and right direction until a collision happens and the one generated by the A$^*$ algorithm. A pair $a = <w_i,p>$ denotes a path primitive, where $w_i\in W$ is the selected subject to actuate with the move path $p= (<x_1,y_1,\phi_1>,<x_2,y_2,\phi_2>,...)$ consisting of the $x, y$ coordinates and the orientation $\phi$.

\textbf{Rewards:} The reward function $r(s,a):(S,A)\rightarrow$ $R$ is constructed from the current observation state $s_{t}$ and action $a_t$, which is shown in Table \ref{tab1}.
\vspace{-0.2cm} 
\begin{table}[!ht]
\begin{center}
\caption{Types of Rewards}
\begin{tabular}{cccc}
\hline
            & \textbf{Reward}  &               & \textbf{Reward} \\ \hline
Move        & -1               & Arrival & 4               \\
Success     & 50               & Leave   & -4              \\ \hline
\end{tabular}
\label{tab1}
\end{center}
\vspace{-0.3cm} 
\end{table}

\textbf{Episodes:} An episode $e_k \in E$ consists of a sequence of states $S$, actions $A$ and rewards $R$. For each episode, the number of time steps $t=1,2,...,T$ are smaller than $T_{max}$.

\textbf{Done:} Only success and failure can lead to an episode done. If the robot pushes all objects from the initial pose to the target pose within $T_{max}$, the task is successful and the episode is done. If the time steps reach $T_{max}$, the task is failed and the episode is also done.

\section{Learning Hierarchical Policy}
\subsection{High-level Policy}
In the high-level policy, the robot focus on \textit{what order to follow}. We propose an MCTS algorithm guided by a policy network which is trained by imitation and reinforcement. The pipeline is shown in Fig \ref{fig3_pipline}.

\subsubsection{Imitation by Supervised Learning}
It will cause the agent to make a lot of trial and error at the beginning of the training that we use unsupervised reinforcement learning to train a policy value network directly. This will lead to a lot of time consumption and make it difficult for the network to converge to a good performance. Thus, we first use supervised learning to make the policy network imitate the actions of experts to make decisions. 

\textbf{Dataset:} The expert $E$ constantly makes an action $a$ according to the state $s$ of the environment, which interacts with the environment to obtain the reward $r$ until the task is done. In \cite{wang2020scene}, the author has trained his own network, and we can directly use this model to interact with the environment as expert data to avoid complicated manual annotation. We can get a dataset including state and action pairs to train our policy network by collecting thousands of episodes. 

\textbf{Training:} In imitation, the states $S$ in the dataset are used as the input of the network, and the actions $A$ are used as the label to supervise the training of the network. Using a policy network to approximate the expert policy function $\pi$: S $\rightarrow$ A. $\pi_{\theta}$ denotes a policy network with $\theta$ being the weights of the network. We show that $\pi_{\theta}(S) = A \approx \pi(S)$, which can be learned with the training dataset that we collected before. After feeding the network with inputs $S_{train}$ and corresponding expert actions $A_{train}$, we apply the cross-entropy loss function $L$, such that
\begin{equation}
    \theta = \mathop{\arg\max}_{\theta} L(A_{train}, \pi_{\theta}(S_{train}))
\end{equation}
Here, we use the Adam optimizer to update the parameter $\theta$.

\subsubsection{Improvement by Reinforcement Learning}
As the number of objects increases, it is difficult for experts to make optimal decisions. However, reinforcement learning can explore a better policy than experts through multiple trials and errors. Thus, it is not enough to imitate experts only. We need to further improve the performance of the policy network through reinforcement learning so that it can exceed experts.

Thus, our goal is to learn an advanced policy network. The policy network can predict the probability of each action being selected based on the state of the environment. We can use the probability distribution to sample the action to interact with the environment. We use categorical distribution because of five discrete path primitives. The function of policy and value is approximated by:
\begin{equation}
    \pi_{\theta}(s,a) = P[a|s,\theta],\quad Q_w(s,a) \approx Q_{\pi_{\theta}}(s,a)
\end{equation}
According to the Proximal Policy Optimization (PPO) algorithm, the network must be updated according to the following combined objective function, which is maximized in each iteration:
\begin{equation}
    L^{C+VF+S}(\theta) = E[L^{C}(\theta) - c_1L^{VF}(\theta) + c_2S[\pi_{\theta}](s)]
\end{equation}
where $c_1$, $c_2$ are coefficients, and $S$ denotes an entropy bonus to ensure sufficient exploration, and $L^{VF}$ and $L^{C}(\theta)$ are loss functions:
\begin{equation}
   L^{VF}(\theta) = (V_{\theta}(s) - V^{target})^2
\end{equation}
\begin{equation}
    L^{C}(\theta) = E[min(r(\theta)A_w, clip(r(\theta), 1-\epsilon, 1+\epsilon)A_w)]
\end{equation}
where $r(\theta) = \frac{\pi_{\theta}(s,a)}{\pi_{\theta_{old}}(s,a)}$ denotes the probability ratio, $A_w(s,a) \approx Q_w(s,a) - V_{\pi_{\theta}}(s)$ is advantage function, $V_{\pi_{\theta}}(s)$ is a value function based on states, $\epsilon$ is a hyperparameter, $\epsilon=0.2$. $clip(r(\theta), 1-\epsilon, 1+\epsilon)$ modifies the surrogate objective by clipping the probability ratio. Here, we use Adam optimizer to update the parameter $\theta$ and $w$. In this way, we use the PPO algorithm to train our policy value network and learn advanced policy beyond experts.

\subsubsection{Guided Monte Carlo Tree Search}
Our high-level policy leverages the trained policy network to guide MCTS to find the most efficient action sequence to achieve the target. For each state, a search tree is built to generate an action. The root node of this tree represents the current state $s$. Each node of this tree represents a state $s$ and stores a value $v_{node}$ which refers to the current estimated value by value network and accumulates rewards in simulation. For the current state, a finished tree needs many iterations with selection, expansion, simulation, backpropagation, which are shown in Fig. \ref{fig3_pipline}. 

\textbf{Selection:} In this step, MCTS uses the policy $\pi_{select}$ to traverse from the root node to a leaf node. $\pi_{select}$ select the max value node $\hat{s}$ based UCB:
\begin{equation}
   \hat{s} = \mathop{\arg\max}_{s} (C\sqrt{\frac{log(N(s_p))}{1+N(s)}})
\end{equation}
where $N(s)$ and $N(s_p)$ are the numbers of times of visits of the node representing current state $s$ and the parent node representing state $s_p$. $C$ is a coefficient. 

\textbf{Expansion:} If the state of the selected node is not the target, we expand a child node by $\pi_{net}$ policy and store the new state. More specifically, we use the policy $\pi_{net}(\hat{s},a)$ estimated by the policy network to predict action $\hat{a}$ by taking the selected node which representing state $\hat{s}$ as input. 
\begin{equation}
   \hat{a} \sim \pi_{net}(\hat{s},a)
\end{equation}
After that, we get the expended node related to a state $s^{(0)}$, where $s^{(0)} = e(\hat{s},\hat{a})$. The superscript 0 means the initial layout in simulation.

\begin{figure}[htbp]
    \centerline{\includegraphics[width=0.45\textwidth]{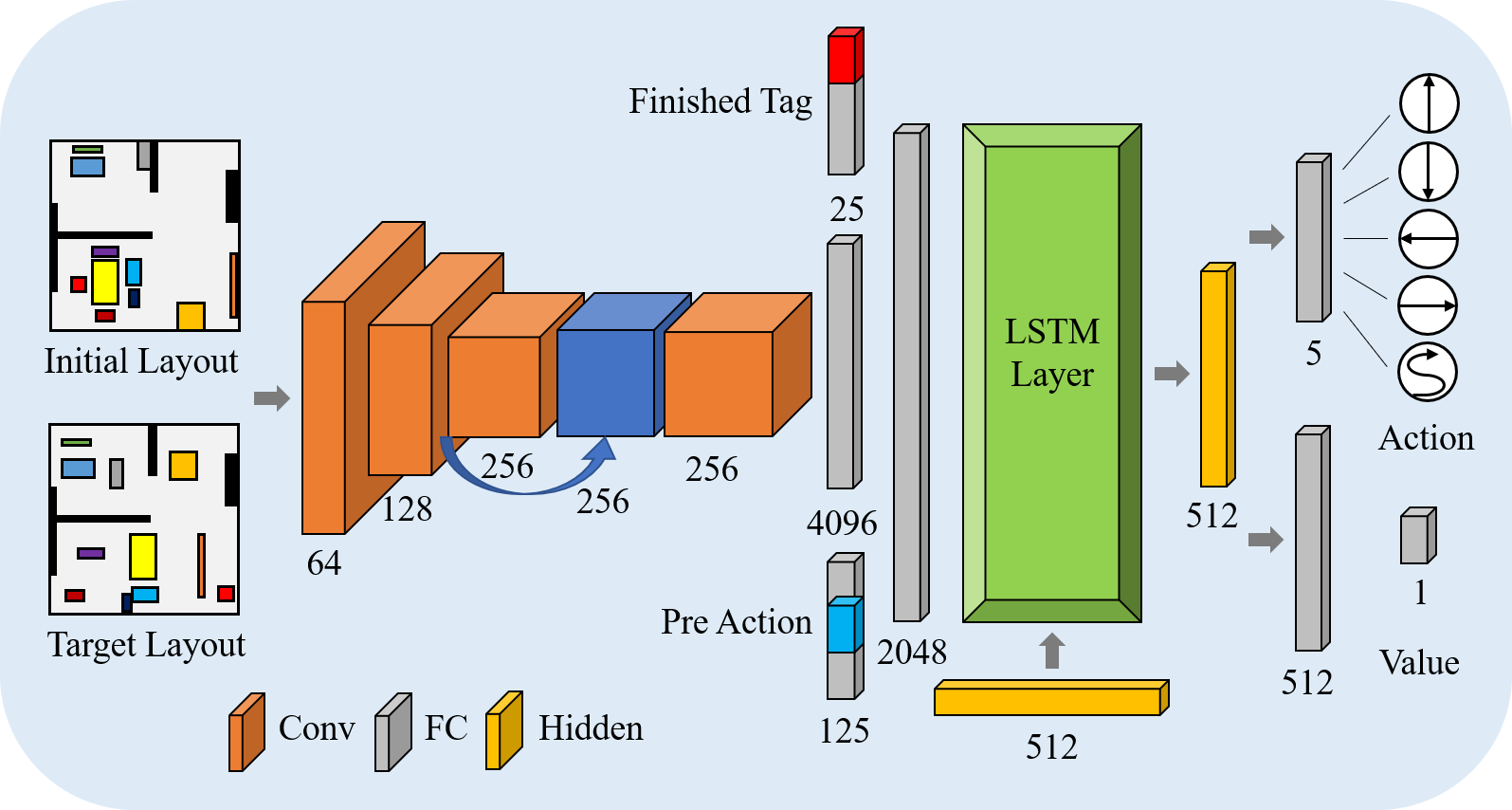}}
    \caption{Illustration of the policy value neural network.}
    \label{fig4_net}
\vspace{-0.5cm} 
\end{figure}

\textbf{Simulation:} After expansion, a simulation is
performed. A sequence of move actions $\{a^{(0)}, a^{(1)}, ...\}$ 
\begin{equation}
   \hat{a}^{(i)} \sim \pi_{net}(s^{(i)},a)
\end{equation}
is chosen according to the policy $\pi_{net}$ from the policy network until the target state or maximum step is reached. 

The node value about $s^{(0)}$ is calculated by the accumulated reward during the simulation.
\begin{equation}
   V_n(s^{(0)}) = \sum_{i=0}^T\gamma^i r(s^{(i)},a^{(i)})
\end{equation}
where $T$ is the simulation steps. The simulation stops if the target or the maximum number of steps is achieved.

\textbf{Back Propagation:} Use the value generated by the simulation to back up and update all the ancestor nodes along the path from the expended node to the root node. For a node related to the state $s$, its estimated node value is updated by:
\begin{equation}
    a^* = \mathop{\arg\max}_{a} V_n(e(s,a))
\end{equation}
\begin{equation}
    V_n(s) = \gamma V_n(e(s,a^*)) + r(s, a^*)
\end{equation}
\begin{equation}
    N(s) = N(s) + 1
\end{equation}
After four steps of iterating for several rounds, a tree with the current state $s_t$ as the root node grew up at the $t-th$ iteration. Our high-level policy uses the grown tree to choose the best action $a_t^*$ in the current state $s_t$.
\begin{equation}
    a_t^* = \mathop{\arg\max}_{a} V_n(e(s,a))
\end{equation}

\subsubsection{The Architecture of the Policy Value Network}
We use a convolutional neural network (CNN) to build a policy value network with local shared parameters to predict the policy and the value. The specific structure is shown in Fig. \ref{fig4_net}. The input is the initial and target state of the object encoded using different channels. The backbone is the residual structure to prevent over-fitting and combine with Long Short Term Memory (LSTM) to embed historical information. Considering the complexity of the NPMP rearrangement problem, we add the finished flag of each object and the previous action to the middle layer. Finally, the output of the LSTM is encoded by a multi-layer perceptron to get the probability distribution vector of action and the predicted value.

\begin{figure*}[ht]
    \centerline{\includegraphics[width=0.98\textwidth]{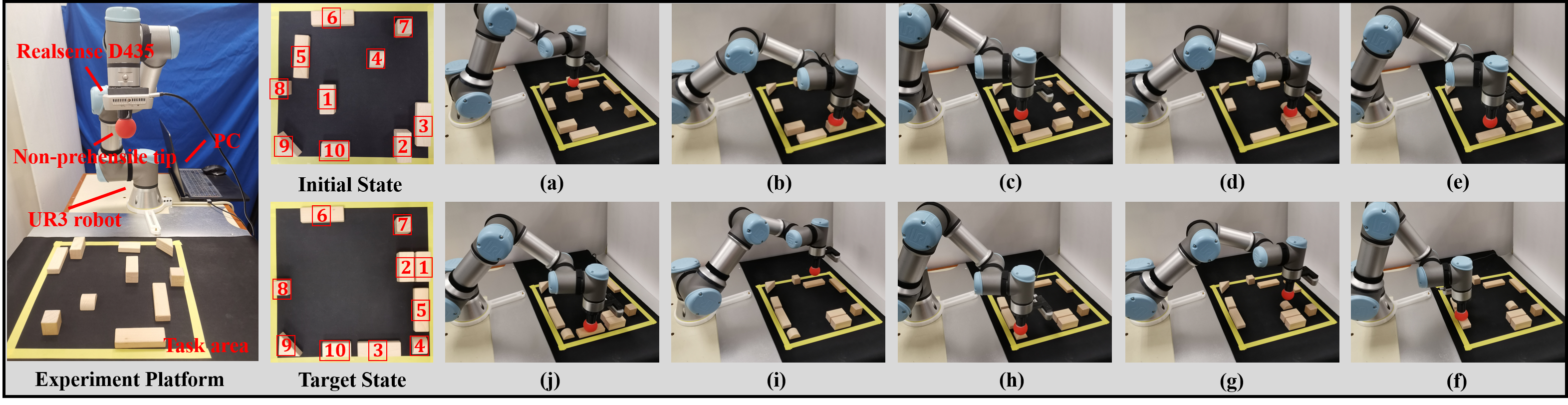}}
    \caption{Qualitative result of algorithm execution in real robot experiments. The sequence of actions is a-j.}
    \label{real_exp}
\vspace{-0.1cm} 
\end{figure*}

\begin{table*}[!ht]
\begin{center}
\caption{Comparing the performance of different methods. SR means Success Rate.}
\begin{tabular}{cccccccccc}
\hline
\multicolumn{2}{c}{Methods}                             & \textbf{PPO} & \textbf{DQN\tiny\cite{wang2020scene}}  & \textbf{IL} & \textbf{PPO+IL} & \textbf{MCTS+Random} & \textbf{MCTS+DQN\tiny\cite{wang2020scene}}  & \textbf{MCTS+PPO} & \textbf{MCTS+PPO+IL(Ours)} \\ \hline
\multirow{3}{*}{5-obj.}  & \multicolumn{1}{c|}{Rewards} & -18.6       &-11.2& 45.8        & 59.2          & -37.5                 & 59.8              & 57.4             & 61.0                \\
                         & \multicolumn{1}{c|}{Steps}   & 34.0        &14.0& 14.2        & 6.4          & 47.9                   & 6.2               & 8.4              & 5.0                 \\
                         & \multicolumn{1}{c|}{SR(\%)}  & 80          &80& 100          & 100         & 20                     & 100               & 100              & 100                 \\ \hline
\multirow{3}{*}{10-obj.} & \multicolumn{1}{c|}{Rewards} & -104.8       &-78.8& 2.5        & 23.3         & -42.6                  & -4.3              & 37.9             & 41.3                \\
                         & \multicolumn{1}{c|}{Steps}   & 50.0        &42.0& 34.9       & 26.3          & 50.0                   & 26.7              & 31.3             & 22.1                \\
                         & \multicolumn{1}{c|}{SR(\%)}  & 0           &20& 50          & 70           & 0                      & 60                & 80               & 70                 \\ \hline
\multirow{3}{*}{15-obj.} & \multicolumn{1}{c|}{Rewards} & -107.2      &-123.2& -35.8     & -6.5           & -31.6                  & -19.0             & 23.3             & 64.1                \\
                         & \multicolumn{1}{c|}{Steps}   & 50.0        &50.0& 50.0       & 44.9          & 50.0                   & 40.6              & 40.7             & 29.9                \\
                         & \multicolumn{1}{c|}{SR(\%)}  & 0           &0& 0           & 20           & 0                      & 30                & 50               & 90                 \\ \hline
\multirow{3}{*}{20-obj.} & \multicolumn{1}{c|}{Rewards} & -110.8      &-125.0& 12.0       & -20.8         & -28.4                  & -26.6             & 21.8             & 51.6                \\
                         & \multicolumn{1}{c|}{Steps}   & 50.0        &50.0& 47.4      & 50.0           & 50.0                   &  50.0             & 46.6             & 43.4                \\
                         & \multicolumn{1}{c|}{SR(\%)}  & 0           &0& 20          & 0            & 0                      & 0                 & 30               & 60                 \\ \hline
\multicolumn{2}{c}{Average Rewards}                     & -85.35      &-84.55& 6.13       & 13.78       & -35.00                    & 2.48             & 35.10             & \textbf{54.50}                 \\
\multicolumn{2}{c}{Average Steps}                       & 46.00       &39.00& 36.63      & 31.90         & 49.48                 & 30.88            & 31.75            & \textbf{25.10}                 \\
\multicolumn{2}{c}{Average SR(\%)}                      & 20.0       &25.0& 42.5        & 47.5         & 5.0                      & 47.5              & 65.0               & \textbf{80.0}              \\ \hline
\label{tab2}
\end{tabular}
\end{center}
\vspace{-0.9cm} 
\end{table*}

\subsection{Low-level Policy}
Search efficiency and policy network are essential in the high-level policy. Thus, in the low-level policy, we design five path primitives (one primitive represents a path, i.e., a sequence of single-step movements) instead of five single-step movements and control the robot to go. This method reduces the width and depth of the search tree to improve search efficiency and focus more on \textit{what order to follow}. It also makes it easier for reinforcement learning to train the policy network to a good performance.

\subsubsection{Path Primitive}
In order to reduce the complexity of search, we design five path primitives to represent the object's action, $\{a_{up}, a_{down}, a_{left}, a_{right}, a_{star}\} \in A$. This path primitive of the low-level policy means moving along the up, down, left, and right direction until a collision happens and the one generated by the A$^*$ algorithm.

\subsubsection{Push to Achieve the Path}
We directly use the original controller of the robot to plan the paths. Compared with the reinforcement learning method which directly applied to the real-time control, our controller makes the robot more safe and stable without abnormal control signals. Since our end effector is operated in a press-slip manner, the control process is greatly simplified.

These path primitive of the low-level policy can be represented as a pair $a = <w_i,p>$, where $w_i\in W$ is the selected subject to actuate with the move path $p= (<x_1,y_1,\phi_1>,<x_2,y_2,\phi_2>,...)$ consisting of path points $x, y$ coordinates and the orientation $\phi$. For each path, we transform them to the robot frame and send the 3D path points to the controller to plan the paths via MoveIt. Each time the high-level policy predicts an action, the low-level policy pushes the selected object to achieve the target path.

\section{Experiments}

\subsection{Experiment Platform and Setup}
We use the simulation\cite{wang2020scene} in our training and test, which can generate various objects in the grid. In training, all algorithms were run on a computer with an NVIDIA 2080Ti GPU. In the real-world configuration, shown in Fig. \ref{real_exp}, we install a Realsense SR300 depth camera which receives RGB and depth information on the wrist of the UR3 robot, after calibration, use the camera to perceive the information of the unstructured environment. Multiple rigid objects of different shapes and sizes are tested on planar work surfaces. We used a laptop with an Intel i7-8750H CPU and NVIDIA 1050Ti GPU to run hierarchical policy, control the real robot and do rearrangement in Kinetic, ROS. 

\begin{figure}[htbp]
    \centerline{\includegraphics[width=0.5\textwidth]{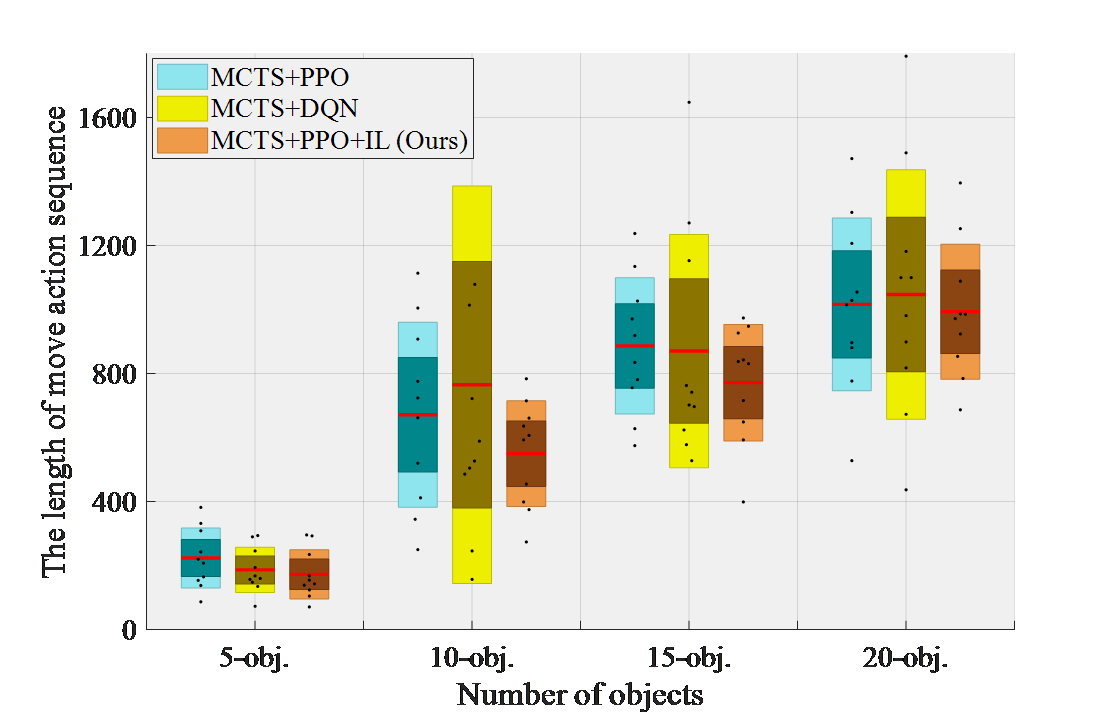}}
    \caption{The length of action sequence of increased objects for MCTS. In each box, black data points are layed over a 1.96 Standard Error of Mean (95\% confidence interval) in dark color and a 1 Standard Deviation in light color. Red lines represent mean values.}
    \label{fig5}
\vspace{-0.3cm} 
\end{figure}

\subsection{Overall Evaluation}

\subsubsection{Comparison with Different Methods}
To compare the performance of our method with others, we tested each method 10 times, and randomly generated 5, 10, 15, 20 objects to reset the environment during each test. Finally, we got the average reward, steps, and success rate, as shown in the Tab. \ref{tab2}. The imitation process used the same data and trained 3K epochs with batch size 64 and learning rate 0.0001. The models in reinforcement learning are trained for 8K iterations with 28 workers, learning rate 0.0002, actor coefficient 1, critic coefficient 0.5, entropy coefficient 0.001 via PPO. The DQN model is the trained same as \cite{wang2020scene}.

We can draw some conclusions from the comparison results. First of all, the method based on MCTS has a higher success rate than other methods, except for MCTS+Random. In the MCTS+Random method, due to the random policy for expansion and simulation, it is difficult to deal with complex sequence decision tasks of multiple objects, so the performance is poor. Secondly, it is obvious to find that imitation learning has excellent effects. If we combine PPO with imitation learning, the average rewards reach 13.78, the success rate reaches 47.5\%, and the steps are only 31.90. Thirdly, we find that combining MCTS with policy net trained by DQN, PPO and ours can achieve fantastic performance, and especially the average rewards reached 54.50, the success rate reached 80.0\% with our MCTS+PPO+IL. Our method exceeded MCTS+DQN used by other papers \cite{wang2020scene} constantly. Finally, as the number of test objects increases, the task becomes more and more difficult. The reward and accuracy become lower, and moving steps increase. Anyway, our method always maintains the best performance and is better able to cope with complex tasks. The detailed comparative experiments are sufficient to prove the effectiveness of our method on the NPMO rearrangement.

\subsubsection{The Effect of Increasing Objects on The Length of Action Sequence for Different Methods}

We compared MCTS+DQN, MCTS+PPO and MCTS+PPO+IL(Ours) the three methods as the number of objects increases, their length of move action sequence change, which is shown in Fig. \ref{fig5}.

As the number of objects increases, the length of the action sequence will increase. In Fig. \ref{fig5}, even if the task becomes more complicated, the mean length of ours is always smaller than others. Our method is more stable, and the distribution of data is more concentrated than other methods. Compared with other methods, our method can stably and excellently cope with tasks of different difficulties.

\begin{figure}[htbp]
    \centerline{\includegraphics[width=0.5\textwidth]{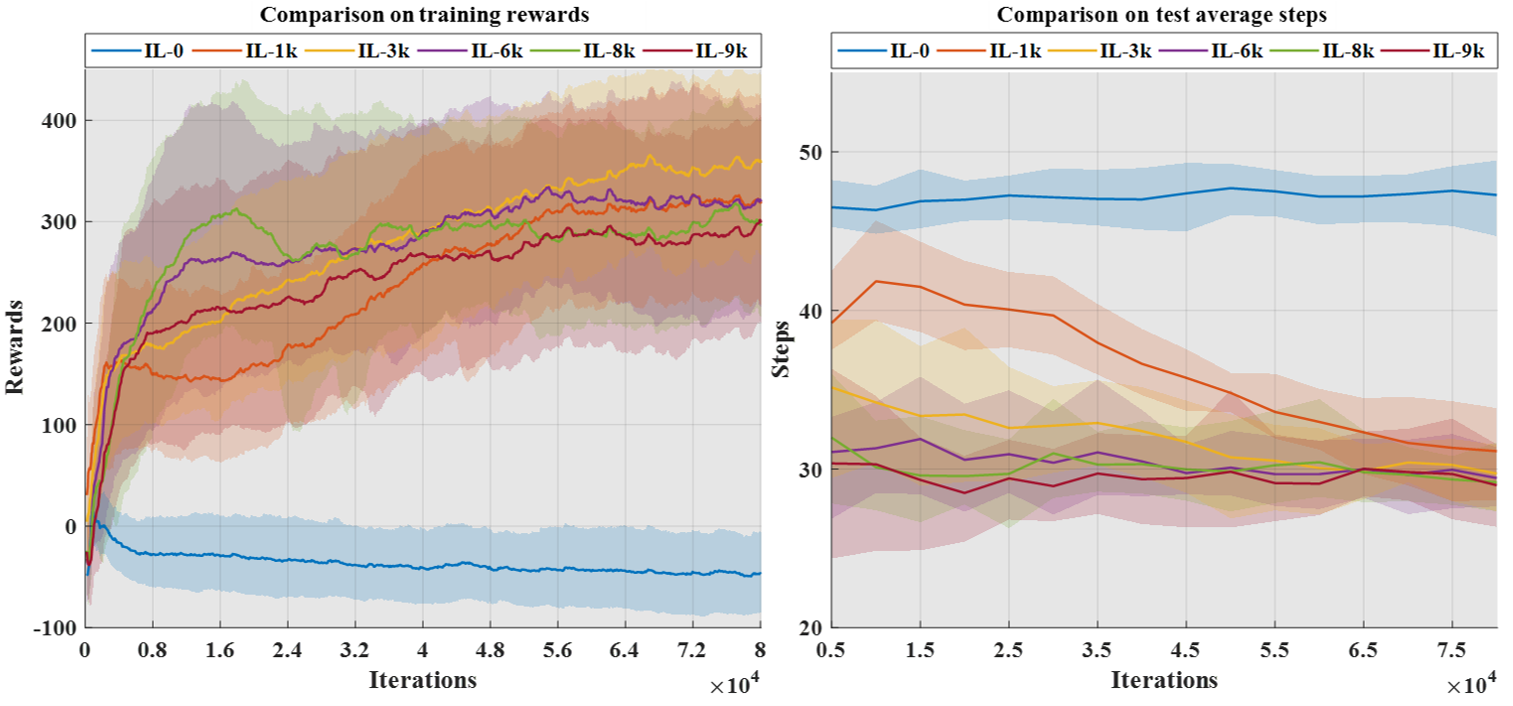}}
    \caption{The impact of different imitation levels on RL.}
    \label{fig7}
\vspace{-0.5cm} 
\end{figure}

\subsubsection{The Impact of Imitation Learning}
To make the policy network converge quickly and reach an expert level first, we use expert data to perform supervised imitation learning on the policy network. The experimental parameter settings of imitation learning are the same as before. We test the impact of different imitation levels (different training epochs: IL-0, IL-1k, IL-3k, IL-6k, IL-8k, IL-9k) on reinforcement learning (RL), which is shown in Fig. \ref{fig7}.

From Fig. \ref{fig7}, the effect is pronounced. If we do not use imitation learning (IL-0) but only use reinforcement learning, the performance is the worst. The training reward is -46.83, and the test average step is 47.27 in convergence. The best model (IL-3k) that combines imitation and reinforcement has the training reward of 357.83, the test average step of 31.13, a reduction of the average step by 16.14. We discovered that with the higher imitation levels, the effect of reinforcement learning may not be better. For example, we found that IL-6k, IL-8k, and IL-9k performed worse than IL-3k. Of course, a good performance cannot be obtained under the too high or too low imitation level. Therefore, we conclude that the appropriate imitation level can help the training of reinforcement learning, but the too high or too low imitation level will result in performance degradation.

\section{Conclusions}

In this paper, we propose a hierarchical policy for the NPMO rearrangement. The low-level policy solves the problem of \textit{how to go}, and the high-level policy solves the problem of \textit{what order to follow}, which improves the performance of the search. After detailed comparative experiments, the excellent performance of our method has been confirmed with a higher success rate, fewer steps, and shorter path length. Our work has important practical significance in scenarios such as service robots, warehousing logistics, and engineering assembly lines. Regarding disadvantages, the MCTS-based method still costs a lot of time, especially with many objects. It is still necessary to further optimize the search algorithm to meet the robot's real-time requirements in future work. It is also a problem worthy of continued study to improve the end effector to make the operation more flexible and adaptable.

\addtolength{\textheight}{2cm}   % This command serves to balance the column lengths

%%%%%%%%%%%%%%%%%%%%%%%%%%%%%%%%%%%%%%%%%%%%%%%%%%%%%%%%%%%%%%%%%%%%%%%%%%%%%%%%

%%%%%%%%%%%%%%%%%%%%%%%%%%%%%%%%%%%%%%%%%%%%%%%%%%%%%%%%%%%%%%%%%%%%%%%%%%%%%%%%

%%%%%%%%%%%%%%%%%%%%%%%%%%%%%%%%%%%%%%%%%%%%%%%%%%%%%%%%%%%%%%%%%%%%%%%%%%%%%%%%

%\section*{APPENDIX}

\section*{ACKNOWLEDGMENT}

This project is supported by Shenzhen Key Laboratory of Robotics Perception and Intelligence (ZDSYS20200810171800001), Hong Kong RGC CRF grant C4063-18G, Hong Kong RGC GRF grant \# 14211420 and Hong Kong RGC GRF grant \# 14200618, which are awarded to Max Q.-H. Meng.


\end{document}